\documentclass[conference]{IEEEtran}
\IEEEoverridecommandlockouts
\usepackage{cite}
\usepackage{amsmath,amssymb,amsfonts}
\usepackage{algorithmic}
\usepackage{graphicx}
\usepackage{multirow}
\usepackage{booktabs}
\usepackage{textcomp}
\usepackage{xcolor}
\def\BibTeX{{\rm B\kern-.05em{\sc i\kern-.025em b}\kern-.08em
    T\kern-.1667em\lower.7ex\hbox{E}\kern-.125emX}}
\begin{document}
\vspace{-0.5cm}
\title{Seq-Masks: Bridging the gap between appearance and gait modeling for video-based person re-identification \thanks{\textbf {Funded by NSFC: 62071292, 61771303 and STCSM 18DZ2270700.}}
}
\vspace{-1.9cm}

\author{
\IEEEauthorblockN{Zhigang Chang\IEEEauthorrefmark{1},  Zhao Yang\IEEEauthorrefmark{1}, Yongbiao Chen\IEEEauthorrefmark{2}, Qin 
Zhou\IEEEauthorrefmark{3}, Shibao Zheng\IEEEauthorrefmark{1}} 
  
\IEEEauthorblockA{\IEEEauthorrefmark{1}Institute of Image Processing and Network Engineering, Shanghai Jiao Tong University, Shanghai, China} 
\IEEEauthorblockA{\IEEEauthorrefmark{3}School of Biomedical Engineering, Shanghai Jiao Tong University, Shanghai, China} 
\IEEEauthorblockA{\IEEEauthorrefmark{2}Shanghai Key Laboratory of Scalable Computing and Systems, Shanghai Jiao Tong University, Shanghai, China} 
}

\maketitle

\begin{abstract}
  Video-based person re-identification (Re-ID) aims to match person images in video sequences captured by disjoint surveillance cameras. Traditional video-based person Re-ID methods focus on exploring appearance information, thus, vulnerable against illumination changes, scene noises, camera parameters, and especially clothes/carrying variations. Gait recognition provides an implicit biometric solution to alleviate the above headache. Nonetheless, it experiences severe performance degeneration as camera view varies. In an attempt to address these problems, in this paper, we propose a framework that utilizes the sequence masks (SeqMasks) in the video to integrate appearance information and gait modeling in a close fashion. Specifically, to sufficiently validate the effectiveness of our method, we build a novel dataset named \textbf{MaskMARS} based on \textbf{MARS}. Comprehensive experiments on our proposed large wild video Re-ID dataset \textbf{MaskMARS} evidenced our extraordinary performance and generalization capability. Validations on the gait recognition metric \textbf{CASIA-B} dataset further demonstrated the capability of our hybrid model. Our codes and dataset \textbf{MaskMARS} will be open-sourced as a strong baseline.
\end{abstract}

\begin{IEEEkeywords}
Multi-modal Fusion, Video-based Person Re-ID, Gait Recognition, appearance model
\end{IEEEkeywords}
\vspace{-0.1cm}
\section{Introduction}
Video-based Person re-identification (Re-ID) ~\cite{gao2018revisiting,liao2018video,zheng2016mars,chen2018video,CHANG2020454, yang2019large} has been drawing proliferating attention from researchers worldwide because of its ubiquitous presence in diversified scenarios ranging from cross-camera object tracking, pedestrian behavior analysis, video surveillance to criminal investigations. Traditional RGB-based, Video-based Person Re-ID methods seeking to build a discriminative appearance model has gained growing momentum in recent years. But the prerequisite is harsh and fatal. It is vulnerable against illumination changes, camera parameters, and scene clutter; the time-effectiveness of the model is short due to clothing change; uniforms and camouflage can easily paralyze the system. Consequently, it is rather reasonable to ponder the possibility of investigating auxiliary information or characteristics.

Gait~\cite{Yu2006A,nambiar2019gait,2018Multi} which manifests the walking style of pedestrians comes to rescue, given its unique advantages ranging from being insensitive to resolution variation to being extremely challenging to impersonate, thus, highlighting the necessity of employing gait feature into performing person re-id tasks~\cite{nambiar2019gait}.
Some gait recognition methods~\cite{Chao2018GaitSet} aim to extract implicit features from videos based on contour or articulated body representations( e.g., UV maps and 2d keypoints), where semantic segmentation (Mask-RCNN~\cite{He_2017_ICCV}) and human pose estimation (DensePose~\cite{G2018DensePose} and OpenPose~\cite{8765346}) can be utilized to model the discriminative representations. However, this stream of works is limited by pedestrian speed, camera perspective, video frame rate, and other factors, leading to low gait recognition performance, especially in practical scenarios. Since video-based person Re-ID methods regard re-identification as sequence matching, which is also challenging due to the random camera view angles and wild scenarios. The goals and settings of these two branches have strong consistency, which naturally leads us to integrate them for better Re-ID performance in wild scenarios.

\begin{figure}[tbp]
\vspace{-0.7cm}
\centerline{\includegraphics[width=9cm]{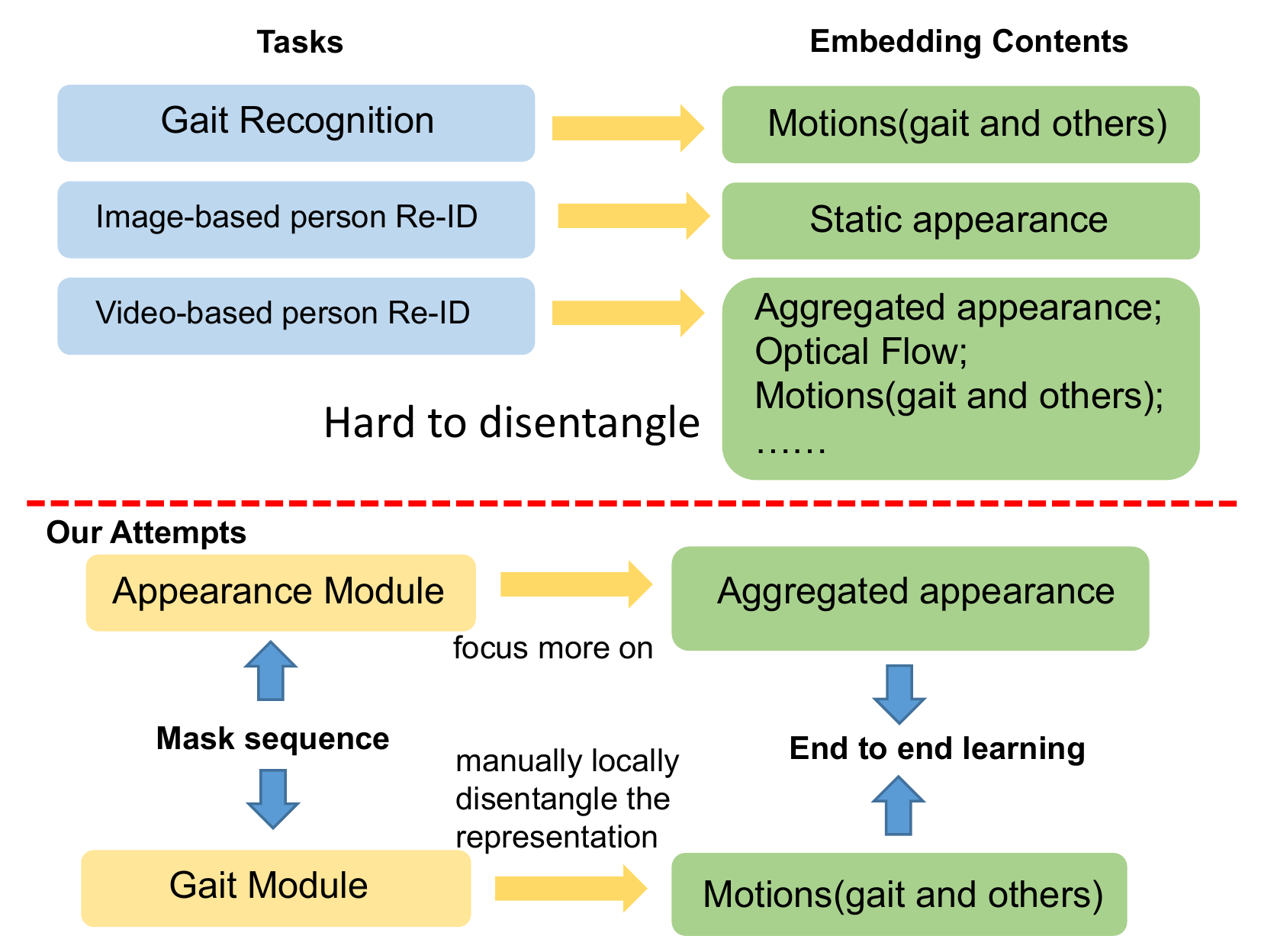}}
\setlength{\abovecaptionskip}{0pt}
\caption{Motivation}
\label{motivation}
\vspace{-0.6cm}
\end{figure}

\begin{figure*}[thp]
\vspace{-1.1cm}
\centerline{\includegraphics[width=18cm]{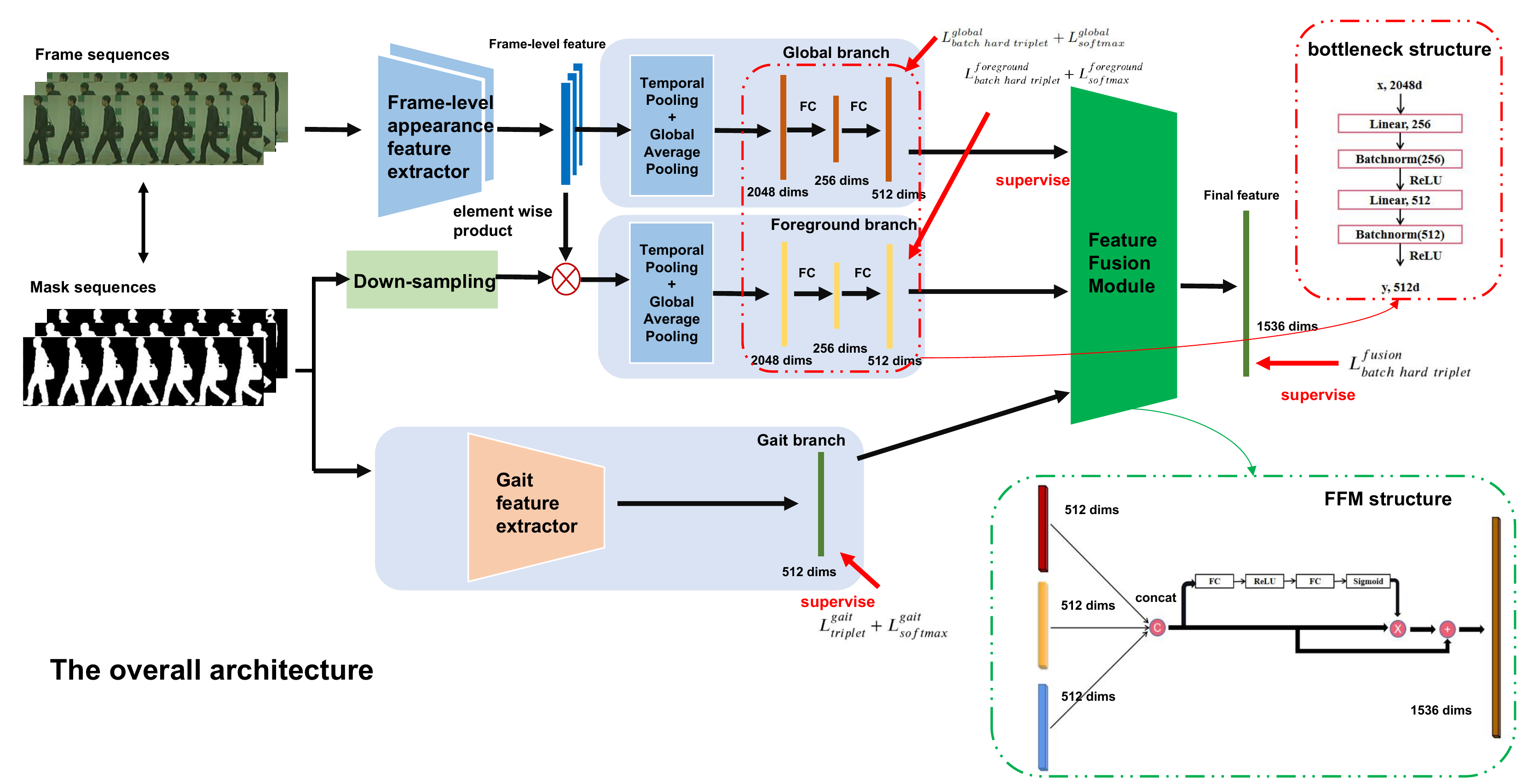}}
\setlength{\abovecaptionskip}{0pt}
\caption{Detailed structure of our Framework}
\label{framework}
\vspace{-0.5cm}
\end{figure*}

In traditional video-based Re-ID, 3D convolution, RNN models, and temporal pooling are commonly used to build an embedding for a video sequence, which integrates multiple information like appearance, motions (gait and other motions), optical flow, etc,. These representations are tightly bound and hard to disentangle. This representation lacks interpretability and may be hard to learn. Our motivation is somehow trying to locally disentangle the gait representation manually by adding a gait branch to a video Re-ID framework. As shown in Fig.~\ref{motivation}, the enhanced gait representation serves as strong prior to joining the final embedding.

In this paper, we utilize the foreground masks of the video sequence to bridge the appearance branch and the gait branch. For the gait branch, we adopt a variant of an advanced gait recognition network Gaitset~\cite{Chao2018GaitSet} where the foreground sequence masks serve as the inputs of the gait branch. For the appearance branch, the foreground sequence masks can be regarded as the saliency map to highlight the foreground human semantic and eliminate background interference. Meanwhile, we also keep the original global branch to reserve global information of the sequence. Finally, we concatenate all the features after each branch and integrate them into a fused representation.

In summary, our contributions are three-fold: First, we proposed a novel end-to-end video person id framework which exploits both appearance and gait information. Second, we built a novel large-scale video Re-id dataset named \textbf{Mask-MASRS}. Third, extensive experiments have demonstrated the validity and effectiveness of our proposed model on both \textbf{Mask-MARKS} and \textbf{CASIA-B} datasets.
\vspace{-0.3cm}

\section{Method}
\vspace{-0.15cm}
The overall network architecture is shown in Fig \ref{framework}. The network contains 3 modules in total: Appearance Module, Gait Module, and Feature Fusion Module. The appearance module is comprised of the global branch and the foreground branch. For each input video sequence (each sequence contains T frames), we use foreground extraction methods (such as Mask RCNN) to extract the pedestrian foreground sequence masks. 

In Appearance Module global branch, the backbone neural network extracts the feature maps of each frame of the sequence; then the global average pooling and temporal average pooling operations flatten and aggregate these feature maps of one sequence to a 2048 dimensions vector, after which a designed bottleneck block reduces the vector to 512 dimensions. As for the foreground branch, the output feature maps of the global branch backbone network are re-used and conducted dot multiplication with the resized foreground masks, which solely average pool the foreground region of the heatmaps and represent the feature of foreground (also 2048 dimensions). Another bottleneck block reduces the foreground vector to 512 dimensions in the same way.

\vspace{-0.1cm}
Meanwhile, Gait Module randomly samples K frames of foreground sequence masks as the inputs of the module. After the Gait Module (a variant of Gaitset~\cite{Chao2018GaitSet}), we obtain a gait feature (a 512-dimension vector) of each sequence masks. 

Finally, the global appearance feature, the foreground appearance feature, and gait feature are concatenated to a 1536-dimension feature vector. We use the Feature Fusion Module to get the fusion feature vector (1536 dimensions). Finally, the fusion feature vector is used for the final representation of the video. We will further introduce the details of each module in the following subsections.
\vspace{-0.2cm}
\subsection{Backbone network and bottleneck structure}
\vspace{-0.1cm}
We adopt ResNet-50~\cite{he2016deep} as the backbone network. By removing the final fully connected layer and changing the last stride from 2 to 1 in $res\_conv5$, we can obtain a larger feature map incorporating richer feature information. For a typical image input size of 256 $\times$ 128, the output of the modified ResNet-50 is 16 $\times$ 8.

Both the global branch and the foreground branch adopt the bottleneck structure. The structure of the bottleneck is depicted in Fig.~\ref{framework} (3), which is comprised of two fully connected layers, each followed by a normalization layer and a ReLU function. Compared to a fully connected layer (2048 $\times$ 512), the bottleneck structure results in notable network parameter reduction.
\vspace{-0.3cm}
\subsection{Gait Branch}
\vspace{-0.1cm}
We adopt and modify an advanced state-of-the-art gait recognition method GaitSet to obtain the gait feature. Different from other template-based and sequence-based methods, GaitSet treats the input as a set of disordered pedestrian contour images. Since pedestrian contours at different time flames exhibit different shapes intuitively, even if the contour sequence is re-shuffled, they can be rearranged into correct orders according to the shapes. The set-based GaitSet method inherently has a processing requirement for sets: sort independence, where the result has nothing to do with the order of input contours. 

\begin{figure}[tbp]
\vspace{-1.3cm}
\centerline{\includegraphics[width=9cm]{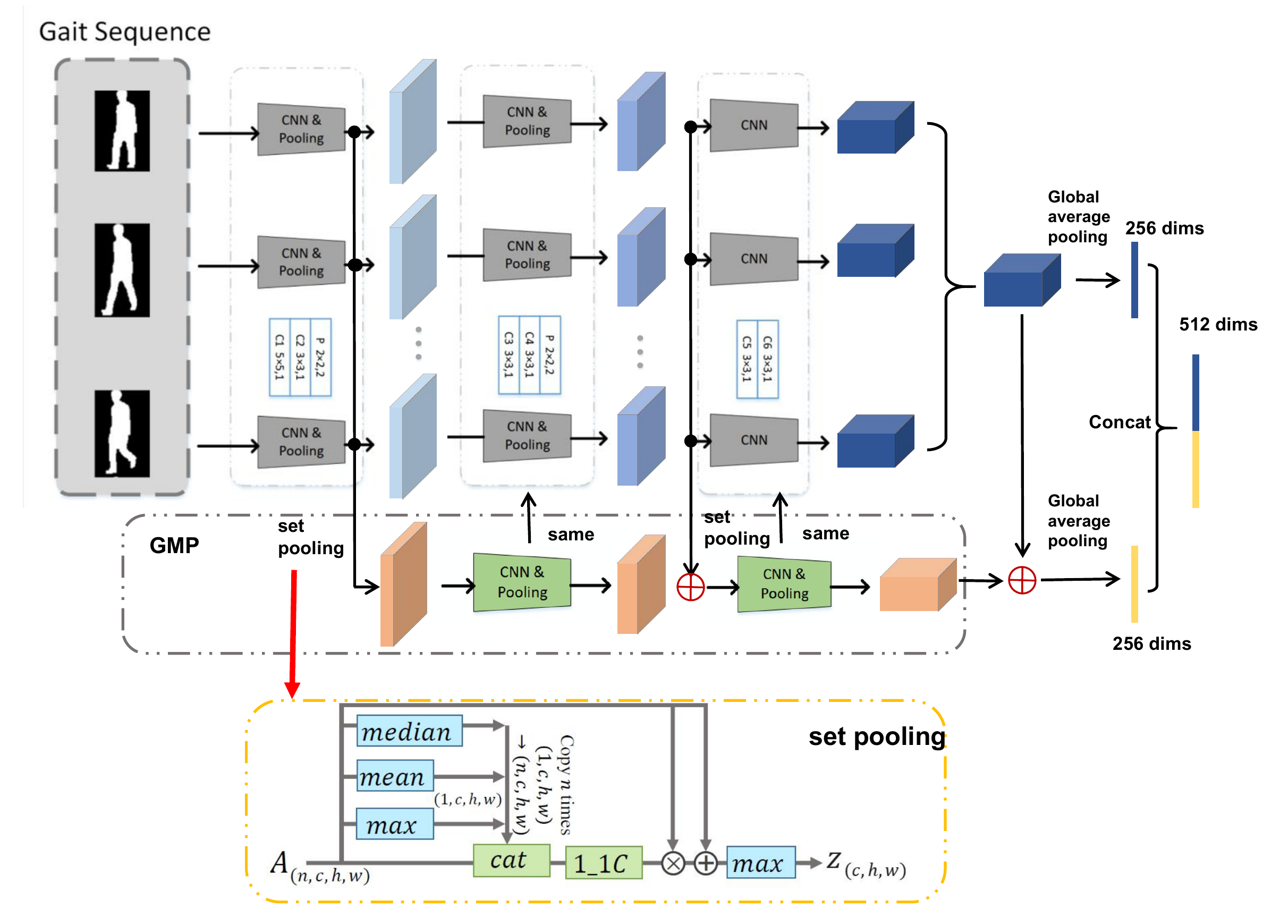}}
\setlength{\abovecaptionskip}{0pt}
\caption{The design of the Gaitset}
\label{Gaitset}
\vspace{-0.6cm}
\end{figure}

Compared with the original GaitSet, the modified GaitSet network reduces the number of horizontal strips at the end of the network, achieving the purpose of reducing the dimension of feature, and the performance is slightly reduced. The modified GaitSet architecture is shown in Fig.~\ref{Gaitset}. The architecture settings before Global Pooling are the same as the original network (also shown detailly in fig.~\ref{framework}). The Set Pooling (SP) module is used to integrate each frame-level feature to form a set-level feature. As we know, a deeper convolutional layer has a larger receptive field. Shallow feature maps contain more local and fine-grained information, while deep feature maps contain global and coarse-grained information. Therefore, MGP(Multilayer Global Pipeline) is designed for aggregating the set-level features from different layers’ outputs. The MGP structure is consistent with the main branch, with the same convolutional layer and pooling layer design, but the parameters are not shared with the main branch. 

We obtain two 128-dimension vectors from the main branch and MGP after global pooling. After two independent fully connected layers, vectors are mapped to 256 dimensions. During the training phase, these two feature vectors are associated with two separate training losses. In the inference stage, they will be concatenated into a 512-dimensional feature vector.

Set Pooling is designed as Fig.~\ref{Gaitset}. It uses 3 kinds of statistical operations that are independent of order: $max ()$, $mean ()$ and $median ()$. $1\_1C$ represents a 1$\times$1 convolutional layer, which is used to fuse cascaded statistical features. Attention mechanism and residual structure are also designed for aggregation and maintaining the information capacity while accelerating and stabilizing convergence. Global Pooling computes the sum of Global Average Pooling and Global Max Pooling.

\vspace{-0.2cm}
\subsection{Feature Fusion Module (FFM)}
\vspace{-0.1cm}
We refer to the channel attention mechanism designed in SeNet to design a feature fusion network, shown as Fig.~\ref{framework}. The global appearance feature, the foreground appearance feature, and the gait feature are concatenated to a 1536-dimension vector. Then we use the Bottleneck mechanism (ratio=8) to construct a channel attention mechanism to promote the exchange of information between different features. At the same time, the residual structure is used to accelerate and stabilize convergence. Finally, a 1526-dimensional fusion feature vector is obtained.

\vspace{-0.3cm}
\subsection{Loss Functions}
\vspace{-0.1cm}
\label{sec:loss}
During training phase, batch-all triplet~\cite{hermans2017defense}, batch-hard triplet loss~\cite{hermans2017defense} and SoftMax loss with Label-Smoothing Regularization (LSR)~\cite{2015Rethinking} are employed to train the network. The loss of the Appearance Module ($L_{appearance}$) is calculated as shown in Eq.~\ref{eq:loss}; the loss of the Gait Module ($L_{gait}$) and the Feature Fusion Module ($L_{fusion}$) are shown as Eq.~\ref{eq:loss1}.
\begin{equation}
\begin{split}
L_{appearance} 
&= L_{batch\ hard\ triplet}^{global} + L_{softmax}^{global} + L_{batch\ hard\ triplet}^{foreground}\\
&+ L_{softmax}^{foreground}
\label{eq:loss}
\end{split}
\vspace{-0.3cm}
\end{equation}
\vspace{-0.3cm}
\begin{equation}
\vspace{-0.3cm}
\begin{split}
L_{gait} = L_{batch\ all\ triplet}^{gait} + L_{softmax}^{gait}; \ L_{fusion} = L_{batch\ hard\ triplet}^{fusion}
\label{eq:loss1}
\end{split}
\end{equation}

And the total loss is shown as as Eq.~\ref{eq:loss2}.
\begin{equation}
\begin{split}
L_{total} = \lambda_{1} L_{fusion} + \lambda_{2}  L_{appearance} + \lambda_{3}  L_{gait}
\label{eq:loss2}
\end{split}
\end{equation}

\section{Datasets and Data preprocessing}

In this paper, both sequence images and foreground sequence masks are needed as inputs for experiments. To obtain data in wild scenarios, we build a dataset named Mask-MARS based on the MARS dataset~\cite{zheng2016mars}, a large-scale dataset for video-based person Re-ID. We also conduct experiments on the CASIA-B dataset~\cite{Yu2006A}.
\begin{list}{\labelitemi}{\leftmargin=1em}
    \setlength{\topmargin}{0pt}
    \setlength{\itemsep}{0em}
    \setlength{\parskip}{0pt}
    \setlength{\parsep}{0pt}
  \item Mask-MARS: We create the original Mask-MARS dataset by computing the foreground mask of each RGB image in MARS dataset by a strong instance segmentation method. We require a video sequence to contain at least 8 effective foreground masks (not necessarily continuous), where an effective mask is simply defined as the proportion of the foreground area to the original image is not less than 15\%. After screening by these two simple rules, the Mask-MARS data set contains a total of 1250 IDs and 14764 video sequences. The training set contains 624 IDs and 5726 video sequences; the query set contains 626 IDs and 1819 video sequences; the gallery set contains 621 IDs and 7170 video sequences. The length of the video sequence varies from 8 to 920 frames, with an average of 70 frames.
  \item CASIA-B: is a popular gait dataset. CASIA-B~\cite{Yu2006A} has a total of 124 IDs, containing 11 angles (0, 18, 36,..., 180 degrees) and 3 different walking conditions. Walking conditions include normal (NM), each pedestrian contains 6 sequences; carrying bag (BG), each pedestrian contains 2 sequences; wearing a jacket or jacket (CL), each pedestrian contains 2 sequences. so each pedestrian contains 110 sequences. We use the first 74 IDs as the training set and the last 50 IDs as the test set. In the test environment, the first 4 sequences (NM1-4) under NM conditions are retained in the gallery subset, and the remaining 6 sequences are divided into 3 query subsets (NM5-6, BG1-2, and CL1-2).
  
  \end{list}
\vspace{-0.1cm}
We adopt the method in~\cite{2018Multi} to preprocess the foreground mask to achieve alignment. During the experiment, the size of the aligned mask image is set to 64$\times$64. We crop 10 pixels on both the left and right sides of the horizontal direction to obtain the size of 64$\times$44 as the input to the GaitSet network.

In the training phase, we preprocess the color image sequence and its corresponding masks in the following way: (1) Random sequence crop: We set the size of the cropped color image to 256$\times$128, and the corresponding mask size of the output feature map of the last layer of ResNet-50 is 16$\times$8, and the random probability $p = 0.5$. (2) Random sequence flip: we set random probability $p = 0.5$. (3) Image standardization: we use the mean and variance statistics of the 3 channels of the ImageNet dataset to normalize the image.

\vspace{-0.2cm}
\section{Experiments}

\vspace{-0.2cm}
\subsection{Progressive results and analysis}
\vspace{-0.1cm}
For simplicity and clarity, we name the baseline along with other module combines as follows:

\textbf{Appearance Baseline}: the global branch of Appearance Module, training individually, 512-dimension feature (without FFM).
\textbf{Appearance Baseline + Foreground Branch}: whole Appearance Module, joint training, 1024-dimension feature (without FFM).
\textbf{Modified GaitSet}: whole Gait Module, training individually, 512-dimension feature (without FFM).
\textbf{Fusion Network}: our full system, joint end-to-end training, 1536-dimension feature (with FFM), with two version: each pre-trained parts finetuning version and end-to-end trained version (\textbf{end2end}) .

As shown in Tab.~\ref{tablem}, in Mask-MARS, compared with the baseline model, the Rank1 index of the model with the foreground branch increased from 84.7\% to 86.5\%, and the mAP increased from 78.9\% to 80.7\%. Because MARS is a large-scale wild scene dataset with random camera views and large background clutters. Only use the gait branch, the performance is very poor (MAP 10.7\% and rank1 17.7\%). Even if the gait model does not perform well, our Fusion Network still outperforms Baseline + Foreground Branch with a notable margin. The two indicators of Rank1 and mAP have increased by 0.8\% and 3.0\% respectively. Compared to the Baseline, the improvement is even more significant (Rank1 +2.6\% and mAP +4.8\%). 

In the CASIA-B dataset, the Tab.~\ref{table2} records the experimental results of including and excluding the same view sequence as the query in the gallery.  Adding appearance features makes the model more robust to view angle variation. Since the dataset is quite simple, there is basically no background clutter. Therefore, we can see that in the case of NM and BG, the appearance-based model can perform very well. However, when clothes changed (CL case),  the performance of our Fusion Network (end2end) dominates all other models, achieving 72.891\% rank1 accuracy, far exceeding the Modified GaitSet (+19.86\%) and Appearance Baseline (+29.49\%). This reflects that when the appearance of pedestrians changes significantly, the fusion features are often more discriminative than single-modal features. In any case in the CASIA-B dataset (NM, BG, CL), we achieve the best performance compared to a single-modal model.

\begin{table}[t]
\vspace{-1.1cm}
\begin{center}
\caption{Progressive results on Mask-MARS dataset}
\vspace{-0.5cm}
\label{tablem}
\resizebox{9.2cm}{!}{
\begin{tabular}{cccccc}
\hline
{Model}		& {Rank1}	& {Rank5} & {Rank10} &{Rank20} & {MAP}\\
\hline
Appearance Baseline	&84.7  & 94.6 & 95.9& 97.2 & 78.9  \\
Appearance Baseline + Foreground Branch  	& 86.5 & \textbf{95.5} & 96.3 &  97.1 & 80.7 \\
Modified GaitSet           & 17.7 & 34.1 & 43.1 &  51.8 & 10.7 \\
\hline
\textbf{FusionNetwork(finetune)} & \textbf{87.1} & 95.2 & \textbf{96.3} & \textbf{97.2} & \textbf{81.1} \\
\textbf{FusionNetwork(end2end)} & \textbf{87.3} & \textbf{95.6} & \textbf{96.5} & \textbf{97.9} & \textbf{83.7} \\
\hline
\end{tabular}}
\end{center}

\vspace{-0.1cm}
\end{table}

\begin{table}[t]
\large
\vspace{-0.6cm}
\begin{center}
\caption{Comparison of concatenation and fusion effects of different features}
\vspace{-0.5cm}
\label{table2}
\resizebox{9.2cm}{!}{
\begin{tabular}{p{160pt}|ccc|ccc}
\hline
\multirow{2}{*}{Model} &  \multicolumn{3}{c|}{Including the same angle
} & \multicolumn{3}{c}{Excluding the same angle} \\
\cline{2-7} & NM & BG & CL & NM & BG & CL  \\
\hline
\noalign{\smallskip}

AppearanceBaseline	&98.669  & 96.630 & 45.884 &98.536  & 96.475 & 45.400\\
AppearanceBaseline+Foreground Branch  	& 98.405 & 95.960 & 48.843 & 98.245 & 95.729 & 48.364 \\
Modified GaitSet         & 83.570 & 71.284 & 55.934  & 81.964 & 69.168 & 55.036 \\
\hline
\textbf{FusionNetwork(finetune)} &99.455 & 96.782 & 73.634 &99.437 & 96.579 &69.495\\
\textbf{FusionNetwork(end2end)} &\textbf{99.620} & \textbf{96.822} & \textbf{75.950} &\textbf{99.582} & \textbf{96.623} & \textbf{74.891}\\
\hline
\end{tabular}}
\end{center}
\vspace{-0.5cm}
\end{table}



\vspace{-0.2cm}
\subsection{Comparison of concatenation and fusion effects of different features}
\vspace{-0.1cm}
This experiment is to verify the effectiveness of the feature fusion module (FFM). We name models with different setting as follows:
\textbf{GGConcat}: concatenated feature from 2 branches: global branch and gait branch; \textbf{GGFusion}: fusion feature from 2 branches with FFM; \textbf{AGConcat}: concatenated feature from 3 branches: global branch, foreground branch and gait branch without FFM; \textbf{AGFusion}: fusion feature from 3 branches with FFM. The experimental results on the Mask-MARS and CASIA-B datasets are shown in Tab. \ref{table3}. 

\vspace{-0.2cm}
\subsection{Compared with other advanced video-based ReID methods}
\vspace{-0.1cm}
We also reproduce some recent advanced algorithms ~\cite{liao2018video,li2018diversity,chen2018video} to validate the effectiveness of our hubrid model on the dataset we created in Tab. \ref{table4}. The performance of our method is far superior compared to methods that model only the appearance features.
\vspace{-0.1cm}

\begin{table}[th!]
\vspace{-0.4cm}
\begin{center}
\caption{Comparison of concatenation and fusion effects of different features}
\vspace{-0.5cm}
\label{table3}
\resizebox{9.2cm}{!}{
\begin{tabular}{l|cccc|ccc}
\hline
\multirow{2}{*}{Model} &  \multicolumn{4}{c|}{Mask-MARS} & \multicolumn{3}{c}{CASIA-B} \\
\cline{2-8} & rank1 & rank5 & rank10 & mAP & NM & BG & CL  \\
\hline
\noalign{\smallskip}

AGConcat  &86.6  & 95.2 & 96.4  & 81.0 & 99.736  & 97.905 & 68.785 \\
AGFusion  &87.1  & 95.2 & 96.3  & 81.1 & 99.620  & 96.822 & 75.950  \\
GGConcat  &87.0 & 95.4 & 96.5   &81.0 & 99.810 & 97.806 & 75.099\\
GGFusion &87.3 & 95.3 & 96.2   &80.7 & 99.149 & 96.740 & 75.992 \\

\hline
\end{tabular}}
\end{center}
\vspace{-0.3cm}
\end{table}

\begin{table}[ht]
\vspace{-0.5cm}
\centering
\caption{Comparison with the state-of-the-art video-based Re-ID methods on MASK-MARS dataset. }
\vspace{-0.1cm}
\begin{tabular}{cccccc}
\toprule
{Model}		& {Rank1}	& {Rank5} & {Rank10} &{Rank20} & {MAP}\\
\midrule
Non-local+C3D~\cite{liao2018video} & 84.1 & 94.5 & 96.0 & 97.3 &77.2 \\
STAN~\cite{li2018diversity}           & 82.3 & 92.9 & 94.6 & 96.8 &65.7 \\
Snipped~\cite{chen2018video}        & 81.2 & 92.1 & 94.6 &  96.5 &69.4 \\
Snipped+$OF$~\cite{chen2018video}    & 86.3 & 94.7 & 95.7 & \textbf{98.2} &76.1 \\
\bottomrule
\textbf{Ours (end2end)} & \textbf{87.3} & \textbf{95.6} & \textbf{96.5} & 97.9 & \textbf{83.7}\\
\hline
\end{tabular}
\label{table4}
\end{table}

\vspace{-0.4cm}
\section{Conclusion}
\vspace{-0.2cm}
\label{sec:refs}
This paper propose an end-to-end framework which utilizes the sequence masks (SeqMasks) in each video to jointly exploit the power of appearance and gait in video Re-ID. Experiments on Mask-MARS dataset evidence the favorable performance and generalization ability of the proposed algorithm. Further validations on gait recognition metric CASIA-B dataset highlight the performance of our hybrid model.

\bibliographystyle{IEEEbib}
\bibliography{czg}

\end{document}